# Enhancing Classification Performance via Reinforcement Learning for Feature Selection

Younes Ghazagh Jahed
Dept. of Electrical Engineering,
faculty of engineering and technology
University of Mazandaran
Babolsar, Mazandaran Province, Iran
y.ghazagh09@umail.umz.ac.ir

Seyyed Ali Sadat Tavana
Dept. of Computer Engineering,
faculty of engineering and technology
University of Mazandaran
Babolsar, Mazandaran Province, Iran
s.sadat08@umail.umz.ac.ir

*Abstract—* **Feature selection plays a crucial role in improving predictive accuracy by identifying relevant features while filtering out irrelevant ones. This study investigates the importance of effective feature selection in enhancing the performance of classification models. By employing reinforcement learning (RL) algorithms, specifically Q-learning (QL) and SARSA learning, this paper addresses the feature selection challenge. Using the Breast Cancer Coimbra dataset (BCCDS) and three normalization methods (Min-Max, l1, and l2), the study evaluates the performance of these algorithms. Results show that QL@Min-Max and SARSA@l2 achieve the highest classification accuracies, reaching 87% and 88%, respectively. This highlights the effectiveness of RL-based feature selection methods in optimizing classification tasks, contributing to improved model accuracy and efficiency.**

## I. Introduction

Effective feature selection is essential for classification tasks, enabling the identification of key features that substantially enhance the predictive accuracy of models [1], [2]. By efficiently filtering out irrelevant or redundant features, feature selection not only improves the accuracy and efficiency of the classification models but also reduces overfitting, simplifies model interpretation, and enhances generalization to new data [3], [4]. Moreover, feature selection can lead to faster model training and reduced computational costs, making it essential in addressing the high-dimensional and complex nature of real-world datasets.

In this context, subsequent studies delve into various approaches for optimizing the feature selection process. For instance, in [5], the study frames feature selection as a reinforcement learning problem using temporal difference method and optimal graph search to navigate through the state space of feature subsets efficiently. By introducing a low-cost evaluation function and exploring feature sets systematically, the proposed approach demonstrates strong performance in comparison to existing feature selection strategies across various datasets. The authors in [6], introduce a novel feature selection method using multi-agent reinforcement learning with main and guide agents to select optimal features based on rewards. By comparing the behavior of main agents to guide agents and updating Q-values accordingly, effective feature subsets are efficiently identified for classification tasks, as demonstrated through improved classification accuracy on various datasets. Reference [7] introduces PA-FEAT, a fast feature selection method using multi-task deep reinforcement learning to generalize feature selection knowledge. PA-FEAT simplifies the selection process by employing a single agent with a restructured choice strategy, improving efficiency and reducing computational costs. In [8] discusses the importance of feature selection methods in machine learning to reduce computation time, improve prediction performance, and enhance data understanding. It provides an overview of Filter, Wrapper, and Embedded methods, highlighting their applicability in various machine learning problems through experiments on standard datasets. The authors in [9], introduce a novel approach for feature selection in reinforcement learning by utilizing dynamic Bayesian networks to deduce minimal feature sets and improve performance in tasks such as stock trading. The experiments conducted show that this method effectively reduces computational expenses while enhancing planning outcomes. Authors in [10], highlight the importance of feature selection in bioinformatics applications and discuss the various techniques available. It provides a taxonomy of feature selection methods and explores their potential in both established and emerging bioinformatics tasks. Reference [11] introduces a feature selection method using multi-agent reinforcement learning to choose optimal gene subsets for classification in gene expression datasets. Each feature is represented by an agent that decides its selection based on obtained rewards, updating Q-values to determine the effectiveness of features. Results on various datasets demonstrate significant accuracy improvements, showcasing the method's capability to select valuable features and enhance classification accuracy. In [12], the Dynamic Key Feature Selection Network (DKFSN) proposes a reinforcement learning framework for accurate feature selection in behavior recognition tasks based on time-series sequential data. It uses a baseline network for feature extraction, a dynamic feature selection network to choose key features, and a reward function for training, leading to improved recognition accuracy by eliminating redundant and interfering features. Authors in [13], address the challenge of balancing effectiveness and efficiency in automated feature selection, proposing a novel approach that combines interactive reinforcement learning with decision tree feedback. By leveraging a structured feature hierarchy and personalized reward schemes based on decision tree insights, the method aims to enhance the accuracy and efficiency of feature selection in behavior recognition tasks. Reference [14] introduces a deep reinforcement learning-based network intrusion detection system for securing IoT environments, focusing on feature selection methods. It highlights the challenges of structuring and training the DRL model and emphasizes the crucial impact of selecting appropriate hyper-parameters on the effectiveness and accuracy of the IDS. The study seeks to identify optimal hyper-parameter values through theoretical and empirical evaluations, considering varying network settings and countermeasures to enhance overall network performance. Authors in [15], introduce a novel online feature selection framework called D-AFS, which utilizes dual worlds to evaluate feature importance in control processes using deep reinforcement learning. Through experiments on classical control systems, D-AFS was shown to outperform human experts and recent feature selection baselines in generating effective feature combinations correlated with system dynamic models. Reference [16], addresses the challenge of balancing efficiency and effectiveness in automated feature selection by proposing an Interactive Reinforced Feature Selection (IRFS) framework. The IRFS framework leverages dual trainers with different searching strategies to guide agents in navigating the feature space, resulting in improved performance compared to traditional feature selection methods and reinforced selection approaches. Reference [17] introduces a novel feature selection method using multi-agent reinforcement learning with main and guide agents to select optimal features based on rewards. The method reformulates feature selection as a reinforcement learning problem, where each feature is considered as an agent. Various state representations such as statistic description, autoencoder, and graph convolutional network are used, along with a GMM-based generative rectified sampling strategy, to accelerate convergence. Experimental results demonstrate the effectiveness of the proposed approach in feature subset selection.

This paper compares two RL algorithms, Q-learning (QL) and SARSA learning, to address the feature selection challenge. Implemented on the Breast Cancer Coimbra dataset (BCCDS) [18], the study examines three normalization methods: Min-Max (MM) [19], $\ell 1$ [20], and $\ell 2$ normalization [21].

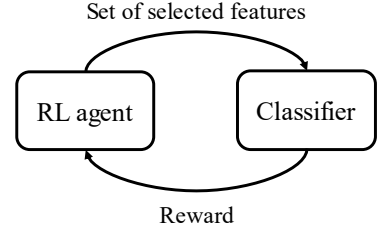

Figure 1. Concept of RL Implementation for Feature Selection.

---

**Algorithm 1.** RL-based Feature Selection
---
Upload database as $\mathbb{D}$
db = $[db_0, db_1, \ldots, db_n]$
**for** $i_{nom} \in [0, n]$ **do:** #preprocessing dataset
$\quad db_{i_{nom}} \leftarrow \Psi_{i_{nom}}(\mathbb{D})$
$\quad$ db $\leftarrow db_{i_{nom}}$
Define $Q_{storage}$
**for** each $db_{i_{nom}}$ in db **do:** #feature selection
$\quad$ **for** $i_{ra} \in [0, r]$ **do:**
$\quad\quad \mathfrak{R}_{i_{ra}}(db_{i_{nom}})$
$\quad\quad Q_{storage} \leftarrow Q_{i_{ra}}$
$\quad\quad$ Extract $\pi^*$
$\quad\quad$ Classify based on $\pi^*$

## II. METHODOLOGY

In this section, the methodology of RL-based feature selection is elucidated. In Figure 1, the concept of RL implementation for feature selection is depicted. Here, the RL agent takes actions, resulting in the selection of a set of features. Subsequently, to evaluate its decisions, the agent sends the chosen action to the classifier, obtaining a reward in return. Algorithm 1 depicts the manner in which $r$ RL algorithms (e.g. QL, SARSA, and etc.) can be utilized to select the optimal features for classifying the given dataset $\mathbb{D}$. Assuming a dataset $\mathbb{D}$ is available, an initial list db is set up to accommodate the normalized versions of $\mathbb{D}$ using a total of $n$ normalization approaches (e.g. Min-Max (MM), $\ell 1$, and etc.). Each normalization approach is iterated through, applying them sequentially to $\mathbb{D}$ using the $\Psi_{i_{nom}}$ function. Here, $\Psi_{i_{nom}}$ represents the function responsible for normalization via the $i_{nom}^{th}$ normalization approach. Following the application of each normalization approach to the $\mathbb{D}$, the resulting new normalized version of $\mathbb{D}$ is saved into $db_{i_{nom}}$ and appended to db. Afterward, the empty set $Q_{storage}$ is initialized. For each $db_{i_{nom}}$ in db and for each $\mathfrak{R}_{i_{ra}}$ in a total of $r$ RL algorithms, $\mathfrak{R}_{i_{ra}}$ is applied to $db_{i_{nom}}$ in order to determine the optimal policy, $\pi^*$ (i.e., selected features), resulting in a total of $n \times r$ $\pi^*$s. Subsequently, based on each $\pi^*$, the dataset $db_{i_{nom}}$ is classified, and the corresponding model is saved for further analysis.

**Reinforcement Learning algorithms:**
RL is a type of machine learning paradigm that focuses on training agents to make sequential decisions in an environment to maximize a cumulative reward [22]–[24]. It draws inspiration from behavioral psychology, where agents learn through trial-and-error interactions with their environment, aiming to discover the most effective actions to achieve desired goals. In this feature selection problem, features are defined as states, with each state having two possible actions: selecting the feature or not selecting the feature. The reward signal sent from the environment to the agent is based on the accuracy of the classifier. If the classifier's accuracy exceeds 70%, the reward is doubled. However, if the agent selects either all or none of the features, it incurs a punishment, resulting in a reduction of the reward by a corresponding punishment value.

| **Algorithm 2**. Q learning |
|---|
| Initialize Q-values arbitrarily |
| Repeat for each episode: |
|   Initialize state s |
|   Repeat for each step in the episode: |
|     Choose action a using an exploration strategy |
|     Take action a, observe reward r and next state s' |
|     Update Q-value for state-action pair: |
|     Q(s, a) = (1 - α) * Q(s, a) + α * [r + γ * max(Q(s', a))] |
|     Update state s to s' |
|   Until terminal state is reached |

- Q-Learning:

QL is a model-free RL algorithm that learns to estimate the value of taking a particular action in a given state [25]. It involves updating a Q-value table iteratively based on the observed rewards and transitions between states. The mathematical formula for updating Q-values can be expressed as:

$$Q(s, a) = (1 - α) * Q(s, a) + α * [r + γ * max(Q(s', a))] \quad (1)$$

where

Q(s, a) (Q-value), representing the expected future reward of taking action 'a' in state 's'.

α (learning rate), determining the weight of new information during Q-value updates.

r, immediate reward received after taking action a in state s.

γ, (discount factor), representing the importance of future rewards.

| **Algorithm 3.** SARSA |
|---|
| Initialize Q-values arbitrarily |
| Repeat for each episode: |
|   Initialize state s |
|   Choose action a using an exploration strategy |
|   Repeat for each step in the episode: |
|     Take action a, observe reward r and next state s' |
|     Choose next action a' |
|     Update Q-value for state-action pair: |
|     Q(s, a) = (1 - α) * Q(s, a) + α * [r + γ * Q(s', a')] |
|     Update state s to s' and action a to a' |
|   Until terminal state is reached |

- SARSA (State-Action-Reward-State-Action):

SARSA is another model-free RL algorithm that learns from experience by directly interacting with the environment [26]. Unlike Q-learning, SARSA updates Q-values based on the action, a' actually taken in the next state s'. The mathematical expression for updating Q-values in SARSA is as follows:

$$Q(s, a) = (1 - α) * Q(s, a) + α * [r + γ * Q(s', a')] \quad (2)$$

## III. RESULTS AND DISCUSSION

In this section, the performance of algorithms applied to the BCCDS using three distinct normalization approaches, namely MM, $\ell 1$, and $\ell 2$, is analyzed. Decision tree [26] was utilized as the classifier. These algorithms were executed within the Google Colab environment, leveraging an Intel Xeon CPU equipped with 2 vCPUs (virtual CPUs) and 13GB of RAM.

Table 1. Summary of BCCDS

| Parameter | Shape/Total Samples |
|---|---|
| $X_{train}$ | (104, 9) |
| $X_{test}$ | (12, 9) |
| $Y_{train}$ | (104, 1) |
| $Y_{test}$ | (12, 1) |
| Class (+) | 52 (44.83%) |
| Class (-) | 64 (55.17%) |

Table 1 provides a summary of the BCCDS. After splitting the dataset into training and testing sets with a ratio of 9:1, the training set comprises 104 data points, while the testing set contains 12 data points. It is important to note that this dataset consists of 9 features, resulting in the shapes of $x_{train}$ and $x_{test}$ being (104,9) and (12,9) respectively. Additionally, each data point is labeled as either positive or negative. Figure 2 illustrates the convergence behavior of algorithms on three normalized datasets. It's worth noting that all RL algorithms are configured with the same parameters: α = 0.03 and γ = 1. Additionally, the agent will receive a punishment of 0.8 in 1000 episodes.

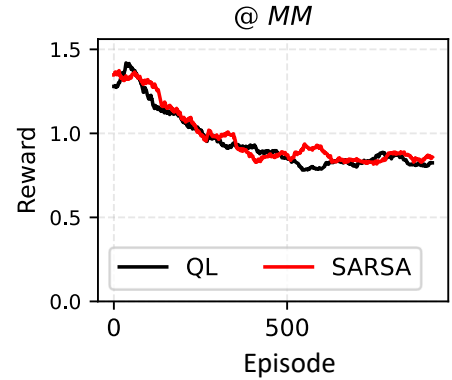

(a)

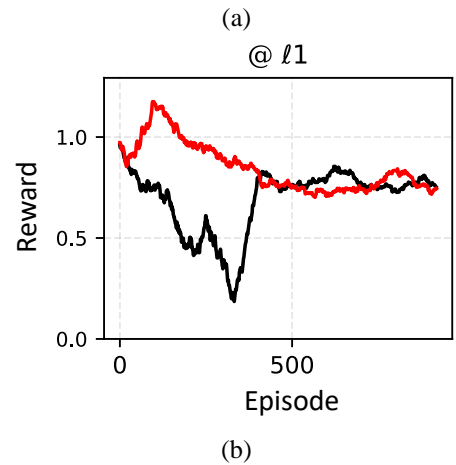

(b)

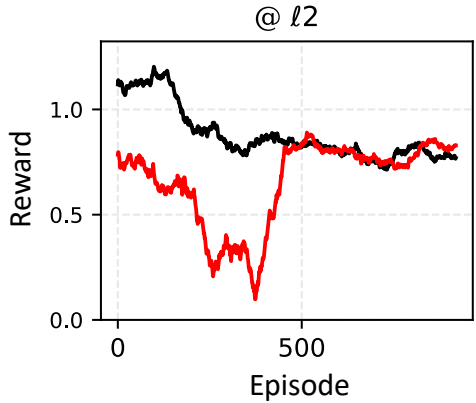

(c)

Figure 2. Convergence Behavior of Algorithms.

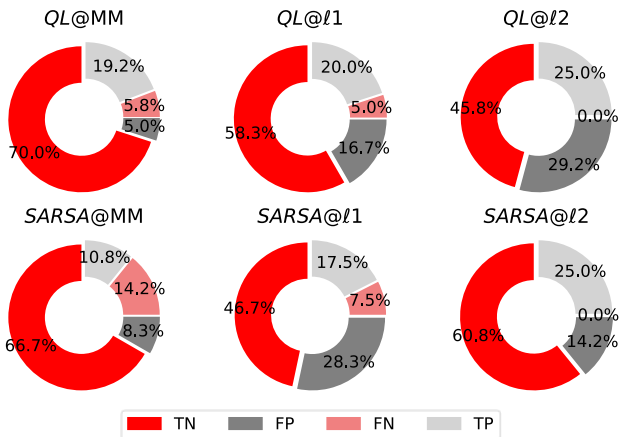

Figure 3. Performance of Classifier with Various Normalization and Feature Selection Methods.

Figure 3 illustrates how the classifier performs under different normalization and feature selection methods. Notably, when utilizing $\ell2$ normalization, no false negatives (FN) are observed. Furthermore, combining feature selection with QL on dataset normalized with MM leads to the lowest occurrences of false positives (FP) and false negatives (FN).

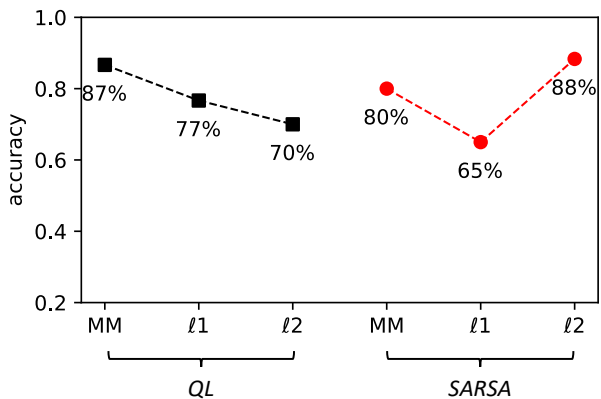

Figure 4. Performance of RL Algorithms on Normalized Datasets

In Figure 4, the accuracy achieved by RL algorithms on the three mentioned normalized datasets is depicted. Notably, QL@MM and SARSA@$\ell2$ exhibit the highest classification performance, with recorded accuracies of 87% and 88%, respectively.

## IV. CONCLUSION

In conclusion, this study delves into the efficacy of RL algorithms, specifically QL and SARSA learning, in tackling the feature selection challenge. Utilizing the BCCDS and evaluating three normalization methods (MM, $\ell1$, and $\ell2$), our findings shed light on the impact of different normalization techniques on classification performance. Notably, employing $\ell2$ normalization results in the absence of FN, indicating its effectiveness in minimizing classification errors. Additionally, the combination of feature selection with QL on datasets normalized with MM demonstrates a significant reduction in both FP and FN. The standout performers, QL@MM and SARSA@$\ell2$, exhibit the highest classification accuracies, reaching 87% and 88%, respectively. These results underscore the potential of RL-based feature selection methods in optimizing classification tasks and improving model accuracy. Moving forward, further exploration and refinement of RL algorithms in feature selection could lead to advancements in classification performance across various datasets and domains.

References

[1] B. Ghojogh *et al.*, "Feature Selection and Feature Extraction in Pattern Analysis: A Literature Review." 2019.

[2] Y. Akhiat, M. Chahhou, and A. Zinedine, "Feature Selection Based on Graph Representation," in *2018 IEEE 5th International Congress on Information Science and Technology (CiSt)*, 2018, pp. 232–237. doi: 10.1109/CIST.2018.8596467.

[3] A. Yassine, C. Mohamed, and A. Zinedine, "Feature selection based on pairwise evalution," in *2017 Intelligent Systems and Computer Vision (ISCV)*, 2017, pp. 1–6. doi: 10.1109/ISACV.2017.8054919.

[4] H. Tan, "Machine Learning Algorithm for Classification," *J. Phys. Conf. Ser.*, vol. 1994, no. 1, p. 12016, Aug. 2021, doi: 10.1088/1742-6596/1994/1/012016.

[5] S. M. Hazrati Fard, A. Hamzeh, and S. Hashemi, "Using reinforcement learning to find an optimal set of features," *Comput. Math. with Appl.*, vol. 66, no. 10, pp. 1892–1904, 2013, doi: https://doi.org/10.1016/j.camwa.2013.06.031.

[6] M. Kim, J. Bae, B. Wang, H. Ko, and J. S. Lim, "Feature Selection Method Using Multi-Agent Reinforcement Learning Based on Guide Agents," *Sensors*, vol. 23, no. 1, 2023, doi: 10.3390/s23010098.

[7] J. Zhang, Z. Luo, Q. Xu, and M. Zhang, "PA-FEAT: Fast Feature Selection for Structured Data via Progress-Aware Multi-Task Deep Reinforcement

Learning," in *2023 IEEE 39th International Conference on Data Engineering (ICDE)*, 2023, pp. 394–407. doi: 10.1109/ICDE55515.2023.00037.

[8] G. Chandrashekar and F. Sahin, "A survey on feature selection methods," *Comput. Electr. Eng.*, vol. 40, no. 1, pp. 16–28, 2014, doi: https://doi.org/10.1016/j.compeleceng.2013.11.024.

[9] M. Kroon and S. Whiteson, "Automatic Feature Selection for Model-Based Reinforcement Learning in Factored MDPs," in *2009 International Conference on Machine Learning and Applications*, 2009, pp. 324–330. doi: 10.1109/ICMLA.2009.71.

[10] Y. Saeys, I. Inza, and P. Larrañaga, "A review of feature selection techniques in bioinformatics," *Bioinformatics*, vol. 23, no. 19, pp. 2507–2517, 2007, doi: 10.1093/bioinformatics/btm344.

[11] M. Kim, J. Bae, and J. S. Lim, "Selecting a Suitable Feature Subset for Classification using Multi-Agent Reinforcement Learning," in *2021 International Conference on Information and Communication Technology Convergence (ICTC)*, 2021, pp. 501–504. doi: 10.1109/ICTC52510.2021.9620934.

[12] T. Zhang, C. Ma, H. Sun, Y. Liang, B. Wang, and Y. Fang, "Behavior recognition research based on reinforcement learning for dynamic key feature selection," in *2022 International Symposium on Advances in Informatics, Electronics and Education (ISAIEE)*, 2022, pp. 230–233. doi: 10.1109/ISAIEE57420.2022.00054.

[13] W. Fan, K. Liu, H. Liu, Y. Ge, H. Xiong, and Y. Fu, "Interactive Reinforcement Learning for Feature Selection With Decision Tree in the Loop," *IEEE Trans. Knowl. Data Eng.*, vol. 35, no. 2, pp. 1624–1636, 2023, doi: 10.1109/TKDE.2021.3102120.

[14] S. Bakhshad, V. Ponnusamy, R. Annur, M. Waqasyz, H. Alasmary, and S. Tux, "Deep Reinforcement Learning based Intrusion Detection System with Feature Selections Method and Optimal Hyper-parameter in IoT Environment," in *2022 International Conference on Computer, Information and Telecommunication Systems (CITS)*, 2022, pp. 1–7. doi: 10.1109/CITS55221.2022.9832976.

[15] J. Wei, Z. Qiu, F. Wang, W. Lin, N. Gui, and W. Gui, "Understanding via Exploration: Discovery of Interpretable Features With Deep Reinforcement Learning," *IEEE Trans. Neural Networks Learn. Syst.*, vol. 35, no. 2, pp. 1696–1707, 2024, doi: 10.1109/TNNLS.2022.3184956.

[16] W. Fan, K. Liu, H. Liu, P. Wang, Y. Ge, and Y. Fu, "AutoFS: Automated Feature Selection via Diversity-Aware Interactive Reinforcement Learning," in *2020 IEEE International Conference on Data Mining (ICDM)*, 2020, pp. 1008–1013. doi: 10.1109/ICDM50108.2020.00117.

[17] K. Liu, Y. Fu, L. Wu, X. Li, C. Aggarwal, and H. Xiong, "Automated Feature Selection: A Reinforcement Learning Perspective," *IEEE Trans. Knowl. Data Eng.*, vol. 35, no. 3, pp. 2272–2284, 2023, doi: 10.1109/TKDE.2021.3115477.

[18] P. J. C. J. M. P. S. R. Patrcio Miguel and F. Caramelo, "Breast Cancer Coimbra." 2018.

[19] Z. Zhao, A. Kleinhans, G. Sandhu, I. Patel, and K. P. Unnikrishnan, "Capsule Networks with Max-Min Normalization." 2019.

[20] B. Shen, B.-D. Liu, Q. Wang, and R. Ji, "Robust nonnegative matrix factorization via L1 norm regularization by multiplicative updating rules," in *2014 IEEE International Conference on Image Processing (ICIP)*, 2014, pp. 5282–5286. doi: 10.1109/ICIP.2014.7026069.

[21] C. Zeng, J. Tian, and Y. Xu, "Analyze the robustness of three NMF algorithms (Robust NMF with L1 norm, L2-1 norm NMF, L2 NMF)." 2023.

[22] X. Qi, D. Chen, Z. Li, and X. Tan, "Back-stepping Experience Replay with Application to Model-free Reinforcement Learning for a Soft Snake Robot." 2024.

[23] X. Ji and G. Li, "Regret-Optimal Model-Free Reinforcement Learning for Discounted MDPs with Short Burn-In Time." 2023.

[24] T. Cai *et al.*, "Model-free Reinforcement Learning with Stochastic Reward Stabilization for Recommender Systems," in *Proceedings of the 46th International ACM SIGIR Conference on Research and Development in Information Retrieval*, in SIGIR '23. ACM, Jul. 2023. doi: 10.1145/3539618.3592022.

[25] C. J. C. H. Watkins and P. Dayan, "Q-learning," *Mach. Learn.*, vol. 8, no. 3, pp. 279–292, 1992, doi: 10.1007/BF00992698.

[26] S. Zou, T. Xu, and Y. Liang, "Finite-Sample Analysis for SARSA with Linear Function Approximation." 2019.